\documentclass[conference]{IEEEtran}
\IEEEoverridecommandlockouts
\usepackage{cite}
\usepackage{amsmath,amssymb,amsfonts}
\usepackage{algorithmic}
\usepackage{graphicx}
\usepackage{textcomp}
\usepackage{xcolor}

\usepackage{booktabs}
\usepackage{multirow}
\usepackage{enumitem}
\usepackage{colortbl}
\usepackage{pifont}
\newcommand{\cmark}{\ding{51}}%

\usepackage[colorlinks,
            linkcolor=blue,
            anchorcolor=blue, 
            citecolor=blue,
            ]{hyperref}

\def\BibTeX{{\rm B\kern-.05em{\sc i\kern-.025em b}\kern-.08em
    T\kern-.1667em\lower.7ex\hbox{E}\kern-.125emX}}

\author{
    \IEEEauthorblockN{
        Aijing Yu$^{1,2}$, 
        Kuangpu Guo$^{4,5}$,  
        Jian Liang$^{3,4}$, 
        and Xiao-Yu Zhang$^{1,2, \ast}$
    }
    \IEEEauthorblockA{
        $^{1}$School of Cyber Security, University of Chinese Academy of Sciences, Beijing, China\\
        $^{2}$Institute of Information Engineering, Chinese Academy of Sciences, Beijing, China\\
        $^{3}$School of Artificial Intelligence, University of Chinese Academy of Sciences, Beijing, China\\
        $^{4}$NLPR \& MAIS, Institute of Automation, Chinese Academy of Sciences, Beijing, China\\
        $^{5}$University of Science and Technology of China, Hefei, China\\
        Email: zhangxiaoyu@iie.ac.cn
    }
    \thanks{$^\ast$Xiao-Yu Zhang is the corresponding author.}
}

\begin{document}

\title{FedBST: Bilateral Self-Training for Unsupervised Federated Classification}

\maketitle

\begin{abstract}
Unsupervised federated learning with Vision-Language Models (e.g., CLIP) offers a promising solution for label-free classification, yet its performance often degrades under severe non-IID shifts.
A pioneering method, FedCoPL, typically relies on self-training strategies that exclusively prioritize high-confidence predictions for pseudo-labeling.
However, this selective approach neglects the information of ambiguous samples (often referred to as  ``waist" samples), leading to the underutilization of data and biased decision boundaries.
To address this limitation, we propose FedBST, a bilateral self-training framework that effectively exploits both high-confidence and ambiguous data.
Specifically, for local optimization, we introduce uncertainty-aware confidence penalization (UACP).
UACP targets ambiguous samples by penalizing overconfident predictions to implicitly promote higher uncertainty, thereby preventing erroneous pseudo-labeling.
At inference, we employ a visual-text ensemble strategy.
This approach aggregates uncertainty-aware local visual prototypes and ensembles them with textual prototypes, achieving lightweight adaptation to local distributions via prototype-based calibration.
Experiments demonstrate that FedBST achieves consistent improvements over existing methods with standard federated aggregation protocols.
\end{abstract} 

\section{Introduction}
\label{sec:intro}
Federated Learning (FL) has established itself as a \textit{de facto} paradigm for privacy-preserving collaborative training, allowing multiple decentralized clients to train a global model without sharing raw private data~\cite{mcmahan2017communication, li2020federated}.
While conventional FL relies heavily on high-quality annotated data~\cite{caldarola2025beyond, liao2025focoop}, obtaining such supervision is often impractical in real-world edge scenarios due to the prohibitive cost of manual labeling and strict privacy constraints~\cite{han2022fedx}.
Consequently, Unsupervised Federated Learning (UFL) has emerged as a critical research direction~\cite{kim2023protofl, yan2024lightweight}.
Early UFL approaches primarily focused on representation learning or clustering~\cite{liao2024rethinking, pourpanah2025federated}, which limits their applicability to direct classification tasks.

Recently, Large-scale Vision-Language Models (VLMs), such as CLIP~\cite{radford2021learning}, have significantly advanced this field by providing rich semantic supervision.
By leveraging the zero-shot capability of VLMs, a pioneering work, FedCoPL~\cite{guo2025cooperative} has explored UFL to Unsupervised
Federated Classification (UFC) settings through cooperative pseudo-labeling, establishing strong baselines. 
This method employs a selective strategy that prioritizes high-confidence predictions as positive pseudo-labels while discarding uncertain ones.

However, this selective paradigm exhibits two limitations under Non-IID shifts.
First, the exclusion of ambiguous samples leads to biased decision boundaries.
By ignoring informative ``waist'' samples near the boundaries, local models tend to overfit to high-confidence easy patterns, failing to learn the robust decision surfaces required for complex Non-IID data.
Second, the exclusive reliance on global textual prompts limits inference adaptability.
Such generic prompts, while semantically correct, lack the granularity to capture client-specific visual variations.
Relying on this global prior neglects the local visual distribution, thereby degrading performance when local data diverges from the generic textual description.

To address these challenges, we introduce \textbf{FedBST}, a Bilateral Self-Training framework designed to maximize data utilization and inference reliability.
Unlike previous methods that rely solely on positive learning from high-confidence data, FedBST employs a bilateral optimization strategy.

For local training, we propose uncertainty-aware confidence penalization (UACP).
Instead of discarding ambiguous samples, UACP treats them as valuable signals for regularization.
Specifically, it explicitly suppresses the maximum predicted probability of these uncertain candidates.
This operation penalizes overconfident predictions, thereby promoting higher predictive entropy to reduce overconfident early assignments to incorrect classes.
For inference, we employ a visual-text ensemble strategy.
This mechanism dynamically aggregates uncertainty-aware local visual prototypes and ensembles them with generic textual prototypes.
This allows the model to leverage the stability of pre-trained textual knowledge while adaptively correcting predictions based on local visual distributions, all without additional parameter optimization.
Our contributions are summarized as follows:

\begin{itemize}
    \item We propose FedBST, a robust UFC framework that integrates bilateral self-training mechanisms. 
    It enables the model to learn simultaneously from high-confidence data while regularizing with ambiguous samples, ensuring robust performance under heterogeneous settings.
    \item We design two complementary components: the uncertainty-aware confidence penalization (UACP) module to regularize decision boundaries using ambiguous data, and a visual-text ensemble (VTE) strategy to bridge generic textual semantics with local visual distributions for reliable inference.
    \item Extensive experiments on multiple benchmarks demonstrate that FedBST outperforms state-of-the-art methods in accuracy and robustness with standard server-side aggregation protocols.
\end{itemize}

\section{Related Work}

\noindent \textbf{Unsupervised Federated Learning.}
Existing UFL approaches primarily focus on representation learning rather than direct classification.
Methods like FedU~\cite{zhuang2021collaborative} and FedEMA~\cite{zhuangdivergence} address heterogeneity via divergence-aware updates, while others employ knowledge distillation~\cite{han2022fedx, kim2023protofl} or clustering~\cite{lubana2022orchestra} to transfer structural knowledge.
However, these representation-focused methods cannot be directly applied to downstream classification tasks without labels.

\noindent \textbf{Federated Learning with VLMs.}
VLMs like CLIP have been integrated into FL to reduce communication costs.
Works such as PromptFL~\cite{guo2023promptfl} and FedCLIP~\cite{lu2023fedclip} optimize prompts but require supervision.
Recent personalization methods~\cite{guo2023pfedprompt, yang2023efficient, li2024global,zhang2026personalized} improve global-local alignment, but still depend on ground-truth, rendering them unsuitable for fully label-free scenarios.

\noindent \textbf{Unsupervised Federated Classification.}
FedCoPL~\cite{guo2025cooperative} recently addressed this challenge by introducing a cooperative pseudo-labeling framework.
It utilizes CLIP's zero-shot predictions to generate pseudo-labels and mitigates label skew through global budget allocation.
As a state-of-the-art baseline in this setting, FedCoPL demonstrates the feasibility of label-free federated classification.
However, it primarily relies on selecting high-confidence samples for training, which inherently neglects the informative value of ambiguous (``waist'') samples.
Building upon this foundation, our work introduces a bilateral self-training strategy to effectively exploit both high-confidence and ambiguous data, further enhancing performance and robustness.

\section{Method}
\label{sec:method}

\subsection{Preliminaries}
\label{subsec:preliminaries}

We consider the UFC setting with $K$ distributed clients. 
Each client $k$ holds a private unlabeled dataset $\mathcal{D}_k$ and shares a pre-trained CLIP backbone comprising a fixed image encoder $E_v(\cdot)$ and a text encoder $E_t(\cdot)$. 
To adapt CLIP to downstream tasks, we introduce learnable visual prompts $\mathbf{P}^v$ and textual prompts $\mathbf{P}^t_k$.
The probability of sample $x$ belonging to class $c$, denoted as $p(y=c|x)$, is computed via the cosine similarity:
\begin{equation} p(y=c|x) = \frac{\exp(\text{sim}(f_v(x), f_t(c)) / \tau)}{\sum_{j=1}^C \exp(\text{sim}(f_v(x), f_t(j)) / \tau)}, \label{eq:prob} \end{equation}
where $f_v$ and $f_t$ are $\ell_2$-normalized visual and textual features extracted using $\mathbf{P}^v$ and $\mathbf{P}^t_k$, respectively. 
$\text{sim}(\cdot, \cdot)$ denotes the cosine similarity (implemented as a dot product for normalized vectors), and $\tau$ is the temperature parameter.
Let $\mathbf{p}_i = [p(y=1|x_i), \dots, p(y=C|x_i)]^\top \in \mathbb{R}^C$ denote the predictive probability vector of sample $x_i$, and $\hat{y}_i = \arg\max_c p(y=c|x_i)$ be the generated pseudo-label.
The goal is to collaboratively optimize these prompts without accessing ground-truth labels.

To mitigate non-IID shifts, a pioneering approach~\cite{guo2025cooperative} utilizes the Cooperative Pseudo-Labeling (CoPL) paradigm, which dynamically updates the training set via two protocols:
\begin{itemize}
    \item \textbf{Global Budget Allocation:} The server calculates a balanced budget $\tilde{U}_k$ for each client to rectify class imbalance.
    \item \textbf{Confidence-Based Selection:} Based on this budget, client $k$ constructs a filtered dataset $\tilde{\mathcal{D}}_k$ by selecting the top-ranked samples with the highest confidence.
\end{itemize}
The standard optimization objective minimizes the cross-entropy loss solely on this filtered high-confidence subset:
\begin{equation}
    \mathcal{L}_{base} = \sum_{k=1}^{K} \mathbb{E}_{(x, \hat{y}) \in \tilde{\mathcal{D}}_k} \ell_{ce}(g(\mathbf{P}^v, \mathbf{P}^t_k; x), \hat{y}).
    \label{eq:loss_base}
\end{equation}

\begin{figure*}[t]
  \centering
  \includegraphics[width=.9\textwidth, trim = 0.5cm 5.5cm 0.5cm 5.5cm,clip]{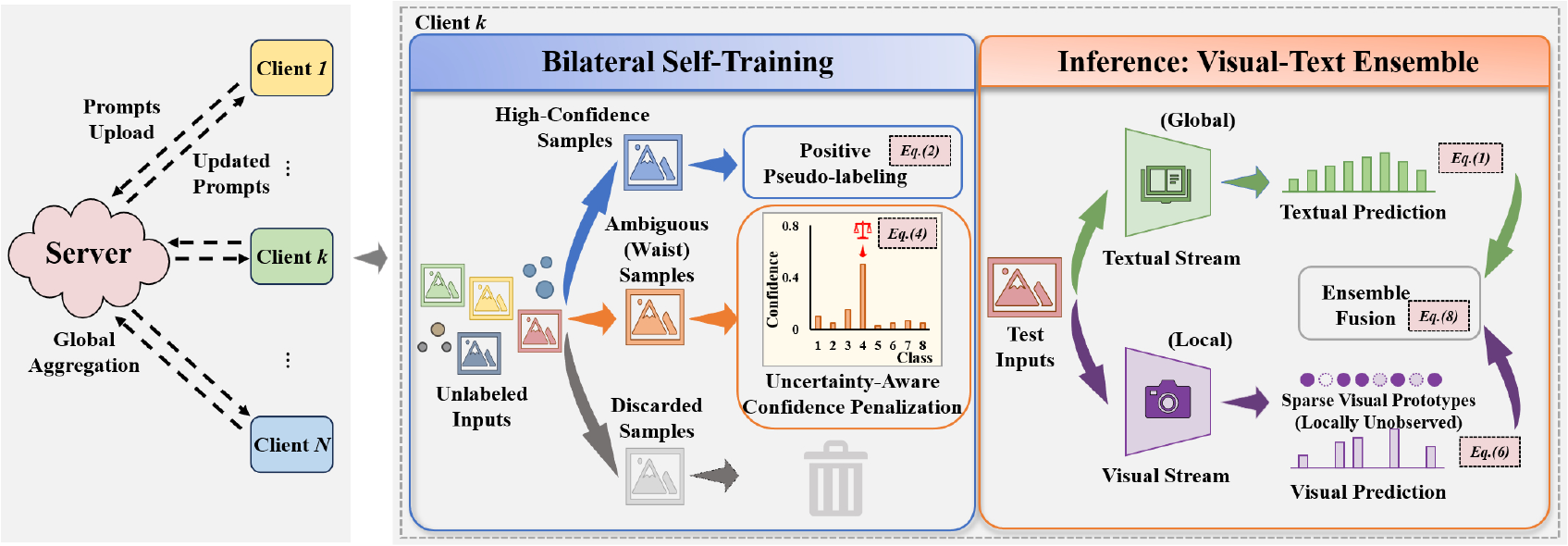}
  \vspace{-8pt}
  \caption{\textbf{Overview of the proposed FedBST framework.} 
  The method consists of two phases:
  \textbf{(Left) Local Bilateral Self-Training:} Rather than using only high-confidence samples (blue), FedBST additionally leverages ambiguous “waist” samples (orange) for regularization. An \textbf{Uncertainty-Aware Confidence Penalization (UACP)} term suppresses the maximum predicted probability of these samples to reduce confirmation bias.
  \textbf{(Right) Visual-Text Ensemble:} At inference, global textual semantics are combined with local visual prototypes. The visual stream is restricted to locally observed classes (hollow dots indicate absent classes), while unobserved classes are handled by the textual stream, ensuring coverage of the full label space.}
  \label{fig:main}
  \vspace{-16pt}
\end{figure*}

While CoPL ensures global balance by prioritizing high-confidence ``head'' samples ($\tilde{\mathcal{D}}_k$), it excludes samples outside the selection budget.
Among them, the ambiguous ``waist'' samples, which lie close to the decision boundary, are not utilized during optimization, despite exhibiting high predictive uncertainty.
As a result, the model lacks explicit constraints on ambiguous predictions, which can encourage overly confident decision boundaries.
To address this, we propose FedBST, which extends the base framework by incorporating an uncertainty-aware confidence penalization (UACP) term applied to these previously ignored samples. The overall framework of FedBST is illustrated in Figure~\ref{fig:main}.

\subsection{Uncertainty-Aware Confidence Penalization}
\label{subsec:uacp}
The budget-based selection strategies discard a substantial portion of local data.
Although these discarded samples are unsuitable for positive pseudo-labeling, they provide information about regions where the model exhibits high predictive uncertainty.
To leverage this signal, we introduce uncertainty-aware confidence penalization (UACP).
Rather than assigning these ambiguous samples to specific classes, which may introduce noise, UACP uses them to regulate model confidence and mitigate confirmation bias.

\noindent \textbf{Identifying Ambiguous Candidates.}
We partition the local dataset $\mathcal{D}_k$ according to the selection protocol.
Let $\tilde{\mathcal{D}}_k$ denote the selected subset of samples with high-confidence positive predictions.
The remaining samples, defined as $\mathcal{R}_k = \mathcal{D}_k \setminus \tilde{\mathcal{D}}_k$, are retained for further consideration, as they may still provide useful regularization signals.
However, $\mathcal{R}_k$ inevitably contains both informative, ambiguous samples and low-quality noise.
To disentangle these two cases, we utilize a dual-filter mechanism to isolate the most informative ``waist'' samples:
\begin{enumerate}
    \item Symmetric Confidence Interval: We filter samples based on the confidence of their tentative predictions. 
    For sample $x_i$, let $\hat{y}_i = \arg\max_c p(y=c|x_i)$ be the predicted class and $\kappa_i = p(y=\hat{y}_i|x_i)$ be the corresponding confidence score. We define a symmetric acceptance interval controlled by a threshold $\tau_{con}$. 
    A sample is retained if $\kappa_i$ lies within $[\tau_{con}, 1-\tau_{con}]$. 
    This excludes both low-confidence noise ($\kappa_i < \tau_{con}$) and high-confidence samples ($\kappa_i > 1-\tau_{con}$) covered by the positive set.
    \item Entropy Verification: To ensure meaningful uncertainty, we further require the predictive entropy to exceed a specific threshold, i.e., $\mathcal{H}(\mathbf{p}_i) > \tau_{ent}$.
\end{enumerate}
The final ambiguous set is defined as:
\begin{equation}
\mathcal{S}_{amb} = \{ x_i \in \mathcal{R}_k \mid \tau_{con} \leq \kappa_i \leq 1-\tau_{con}, \ \mathcal{H}(\mathbf{p}_i) > \tau_{ent} \}.
\end{equation}

\noindent \textbf{Penalization Objective.} 
For the high-confidence set $\tilde{\mathcal{D}}_k$, we retain the standard cross-entropy loss $\mathcal{L}_{base}$ to encourage class separation.
For the ambiguous set $\mathcal{S}_{amb}$, we introduce a confidence penalization term.
As ground-truth labels are unavailable, no class-specific supervision is imposed.
Instead, we penalize overconfident predictions on these uncertain inputs by minimizing the maximum predicted class probability:
\begin{equation}
\mathcal{L}_{uacp} = \frac{1}{|\mathcal{S}_{amb}|} \sum_{x_i \in \mathcal{S}_{amb}} p(y=\hat{y}_i \mid x_i).
\label{eq:uacp_loss}
\end{equation}
Minimizing $\mathcal{L}_{uacp}$ reduces overly peaked predictive distributions on ambiguous samples.
Although this objective does not explicitly compute Shannon entropy, it acts as a lightweight surrogate that discourages overconfident predictions, particularly near decision boundaries.

\noindent \textbf{Total Objective.}
The overall local objective for client $k$ is:
\begin{equation}
\mathcal{L}_{total} = \mathcal{L}_{base} + \lambda \mathcal{L}_{uacp},
\end{equation}
where $\lambda$ acts as a trade-off parameter, balancing discriminative learning on high-confidence samples with explicit confidence control on ambiguous ones.

\noindent \textbf{Global Aggregation.}
Following FedCoPL~\cite{guo2025cooperative}, we adopt partial prompt aggregation.
Specifically, let $\mathbf{P}^v_k$ denote the locally updated visual prompts of client $k$.
The server aggregates these prompts globally as $\mathbf{P}^v_{global} \leftarrow \sum_{k} \frac{n_k}{N} \mathbf{P}^v_k$, while textual prompts $\mathbf{P}^t_k$ remain client-specific to preserve local semantic adaptation.

\subsection{Visual-Text Ensemble}
\label{subsec:ensemble}

To complement the training-time regularization, we employ an ensemble strategy at inference time.
Although global textual prompts encode category-level semantics, they are often insufficient to capture the client-specific visual variations that arise under non-IID local distributions.
Accordingly, we incorporate locally constructed visual prototypes to augment the global textual semantics, enabling effective local adaptation while mitigating overfitting.

\noindent \textbf{Local Visual Prototypes.}
After local training, visual prototypes are constructed from the high-confidence set $\tilde{\mathcal{D}}_k$.
For simplicity and robustness, we aggregate reliable samples to represent the local data distribution.
Specifically, for each class $c$, we sum the visual features of samples predicted as class $c$:
\begin{equation}
\mathbf{v}_c = \sum_{x_i \in \tilde{\mathcal{D}}_k} \mathbb{I}(\hat{y}_i=c) \cdot f_v(x_i).
\end{equation}
The prototype $\mathbf{z}_c$ is obtained by $\ell_2$-normalizing $\mathbf{v}_c$.
To handle non-IID settings, where a client typically observes only a subset of the global categories (i.e., label skew), we utilize a Class Existence Mask $m_c \in \{0,1\}$.
The mask is defined as $m_c = 1$ if $\|\mathbf{v}_c\|_2 > 0$ and $0$ otherwise.

\begin{table*}[t]
\setlength{\tabcolsep}{2pt}
\centering
\caption{Accuracy (\%) of different methods under Dirichlet-based label skews with ViT/B-32 as the backbone. FPL~\cite{menghini2023enhancing}, CPL~\cite{zhang2024candidate}, and CoPL~\cite{guo2025cooperative} are adopted as baseline pseudo labeling (\textbf{PL}) method. \textbf{Bold} and \underline{underline} represent the best and second-best results.}
\vspace{-4pt}
\label{tab: accuracy Dirichlet}
\resizebox{\textwidth}{!}{ 
\begin{tabular}{lccccccccccccccc}
\toprule 
\multirow{2}{*}{Method} & \multirow{2}{*}{\textbf{PL}} & \multicolumn{2}{c}{{DTD}} & \multicolumn{2}{c}{{RESISC45}} & \multicolumn{2}{c}{{CUB}} & \multicolumn{2}{c}{{UCF101}} & \multicolumn{2}{c}{{CIFAR10}} & \multicolumn{2}{c}{{CIFAR100}} & \multicolumn{2}{c}{{Average}} \\ \cmidrule{3-16} 
    & & $\beta = 0.10$ & $\beta = 0.05$ & $\beta = 0.10$ & $\beta = 0.05$& $\beta = 0.10$ & $\beta = 0.05$& $\beta = 0.10$ & $\beta = 0.05$& $\beta = 0.10$ & $\beta = 0.05$ & $\beta = 0.10$ & $\beta = 0.05$  & $\beta = 0.10$ & $\beta = 0.05 $\\ \midrule
CLIP & - & {43.24} &{43.24} & {54.51} & {54.51} & 51.28 & 51.28 & 61.00 & 61.00 & {86.93} &  {86.93} & {64.17} & {64.17} & 60.19 & 60.19 \\ 	
PromptFL~\cite{guo2023promptfl} & FPL & 45.79 & 44.62 & 59.76 & 58.05 & 47.29 & 46.16 & 64.39 & 62.96 & 87.51 & 85.02 & 63.26 & 62.56 & 61.33 & 59.90 \\	
PromptFL~\cite{guo2023promptfl} & CPL & 44.84 & 46.32 & 62.52 & 60.99 & 48.72 & 48.86 & 63.86 & 64.97 & 88.90 & 89.14 & 66.04 & 65.14 & 62.48 & 62.57 \\
Promptprox~\cite{li2020federated} & FPL & 45.15 & 43.88 & 59.36 & 58.59 & 47.04 & 47.01 & 62.91 & 61.27 & 87.13 & 85.62 & 62.77 & 64.12 & 60.73 & 60.08 \\
Promptprox~\cite{li2020federated} & CPL & 43.51 & 45.85 & 59.99 & 62.35 & 49.25 & 48.65 & 64.55 & 63.94 & 90.44 & 90.07 & 65.86 & 66.15 & 62.27 & 62.84 \\	
pFedPrompt~\cite{guo2023pfedprompt} & FPL & 44.56 & 46.19 & 65.95 & 60.52 & 48.48 & 44.42 & 64.37 & 64.18 & 87.45 & 90.72 & 65.08 & 65.83 & 62.65 & 61.98 \\	
pFedPrompt~\cite{guo2023pfedprompt} & CPL & 44.22 & 47.59 & 61.76 & 66.79 & 47.23 & 50.97 & 65.59 & 65.44 & 90.26 & 89.10 & 65.63 & 67.86 & 62.45 & 64.63 \\	
FedOPT~\cite{li2024global} & FPL & 51.97 & 48.71 & 57.76 & 55.62 & 47.99 & 47.05 & 61.83 & 64.56 & 91.32 & 90.73 & 67.51 & 66.50 & 63.06 & 62.20 \\	
FedOPT~\cite{li2024global} & CPL & 37.43 & 39.33 & 51.10 & 48.47 & 46.61 & 45.07 & 58.93 & 57.12 & 91.08 & 91.36 & 57.87 & 59.81 & 57.17 & 56.86 \\
FedCoPL~\cite{guo2025cooperative} & CoPL & \underline{60.89} & \underline{66.37} & \underline{75.76} & \underline{80.26} & \underline{56.09} & \underline{54.80} & \textbf{73.20} & \underline{74.97} & \underline{95.38} & \underline{95.11} & \underline{73.59} & \underline{72.84}  & \underline{72.49} & \underline{74.06} \\ \cmidrule{1-16} \rowcolor[gray]{0.9}
FedBST & UACP & \textbf{61.46} & \textbf{67.77} & \textbf{77.88} & \textbf{81.16} & \textbf{57.25} & \textbf{57.08} & \underline{72.58} & \textbf{75.04} & \textbf{95.47} & \textbf{95.46} & \textbf{75.01} & \textbf{73.92} & \textbf{73.28} & \textbf{75.07} \\
\bottomrule
\end{tabular}
}
\vspace{-10pt}
\end{table*}

\begin{table*}[t]
\setlength{\tabcolsep}{8pt}
\centering
\caption{Accuracy (\%) of different methods under quantity-based label skews with ViT/B-32 as the backbone. FPL~\cite{menghini2023enhancing}, CPL~\cite{zhang2024candidate}, and CoPL~\cite{guo2025cooperative} are adopted as baseline pseudo labeling (\textbf{PL}) method. \textbf{Bold} and \underline{underline} represent the best and second-best results.}
\vspace{-4pt}
\label{tab: accuracy Quantity}
\resizebox{0.8\textwidth}{!}{ 
\begin{tabular}{lcccccccc}
\toprule
{Method} & \textbf{PL} & {{DTD}} & {{RESISC45}} & {{CUB}} & {{UCF101}} & {{CIFAR10}} & {{CIFAR100}} & {Average} \\ \midrule
CLIP & - & {43.24} & {54.51} & 51.28 & {61.00} & {86.93} & {64.17} & 60.19 \\ 
PromptFL~\cite{guo2023promptfl} & FPL & 43.53 & 57.53 & 46.66 & 62.35 & 88.29 & 63.94 & 60.38 \\ 
PromptFL~\cite{guo2023promptfl} & CPL & 43.82 & 61.11 & 47.31 & 64.23 & 88.70 & 67.23 & 62.07 \\
Promptprox~\cite{li2020federated} & FPL & 44.89 & 56.71 & 48.74 & 61.80 & 85.69 & 64.01 & 60.31 \\
Promptprox~\cite{li2020federated} & CPL & 45.53 & 60.77 & 47.61 & 63.86 & {89.21} & 66.49 & 62.25 \\
pFedPrompt~\cite{guo2023pfedprompt} & FPL & 45.09 & 60.46 & 47.71 & 65.39 & 87.54 & 62.82 & 61.50\\
pFedPrompt~\cite{guo2023pfedprompt} & CPL & 45.14 & 66.68 & 50.45 & 65.57 & 87.98 & 65.45 & 63.55 \\
FedOPT~\cite{li2024global} & FPL & 47.65 & 56.04 & 46.61 & 62.95 & 90.74 & 66.76 & 61.79 \\
FedOPT~\cite{li2024global} & CPL & {40.24} & 47.09 & 44.00 & 57.24 & 89.21 & 57.98 & 55.96 \\
FedCoPL~\cite{guo2025cooperative} & CoPL & \underline{56.18} & \underline{81.06} & \underline{56.31} & \underline{72.03} & \underline{94.86} & \underline{73.39}  & \underline{72.31} \\ \cmidrule{1-9} \rowcolor[gray]{0.9}
FedBST & UACP & \textbf{57.63} & \textbf{82.04} & \textbf{56.79} & \textbf{73.01} & \textbf{95.24} & \textbf{74.85} & \textbf{73.26}  \\
\bottomrule
\end{tabular}
}
\vspace{-18pt}
\end{table*}

\noindent \textbf{Ensembled Inference.}
At inference time, predictions are obtained by combining a textual stream with a visual stream.
To ensure the visual stream provides reliable signals, we perform probability normalization strictly within the subset of observed classes.
The two streams are defined as:
\begin{itemize}
    \item Textual Stream: $\mathbf{p}_{text} \in \mathbb{R}^C$, computed via Eq.~\eqref{eq:prob}, serves as a global fallback covering the entire label space.
    \item Visual Stream: We compute the similarity between the test sample and local prototypes. 
    We apply the mask $m_c$ within the Softmax formulation to restrict the probability distribution to locally existing classes:
\end{itemize}
\begin{equation}
p_{visual}(y=c|x) = \frac{m_c \cdot \exp(\text{sim}(f_v(x), \mathbf{z}_c) / \tau_v)}{\sum_{j=1}^C m_j \cdot \exp(\text{sim}(f_v(x), \mathbf{z}_j) / \tau_v)}.
\label{eq:p_visual}
\end{equation}
This formulation ensures that for missing classes ($m_c=0$), the visual probability is zero, while for existing classes, the probabilities sum to 1 (i.e., locally normalized).

The final prediction is obtained by a linear combination of the two streams:
\begin{equation}
\mathbf{p}_{final} = \alpha \cdot \mathbf{p}_{text} + (1 - \alpha) \cdot \mathbf{p}_{visual}.
\end{equation}
Through this ensemble, predictions for locally observed classes are refined by specific visual prototypes, while predictions for missing classes implicitly rely on the global textual knowledge (scaled by $\alpha$), ensuring the model remains functional across the full label space.

\section{Experiments}
\label{sec:exper}

\subsection{Experimental Setup}

\noindent \textbf{Datasets and Baselines.}
We conduct evaluations on six benchmarks: DTD~\cite{cimpoi2014describing}, RESISC45~\cite{cheng2017remote}, UCF101~\cite{soomro2012ucf101}, CUB~\cite{wah2011caltech}, CIFAR-10, and CIFAR-100~\cite{krizhevsky2009learning}.
Following~\cite{guo2025cooperative}, we simulate two non-IID scenarios: (1) \textbf{Dirichlet-based skew}, where class distributions are sampled from a Dirichlet distribution ($\beta$); and (2) \textbf{Quantity-based skew}, where samples are grouped by class and distributed via heterogeneous shards.
We compare with state-of-the-art methods including FedCoPL~\cite{guo2025cooperative}, FedOPT~\cite{li2024global}, pFedPrompt~\cite{guo2023pfedprompt}, PromptProx~\cite{li2020federated}, and PromptFL~\cite{guo2023promptfl}.
Supervised baselines are combined with pseudo-labeling strategies FPL~\cite{menghini2023enhancing} or CPL~\cite{zhang2024candidate} for unsupervised adaptation.

\noindent \textbf{Implementation Details.}
\label{Implementation details}
We use a frozen CLIP (ViT-B/32 by default).
Prompts are optimized via SGD (lr=0.1, momentum=0.9) with cosine decay.
We report the mean accuracy of three trials.
Default hyperparameters are set as $\tau_{\text{con}} = 0.2$, $\tau_{\text{ent}} = 0.5$, $\lambda = 0.001$, and $\alpha = 0.8$.
Other settings are the same as FedCoPL~\cite{guo2025cooperative}.

\begin{figure*}[t]
  \centering
  \includegraphics[width=0.9\textwidth]{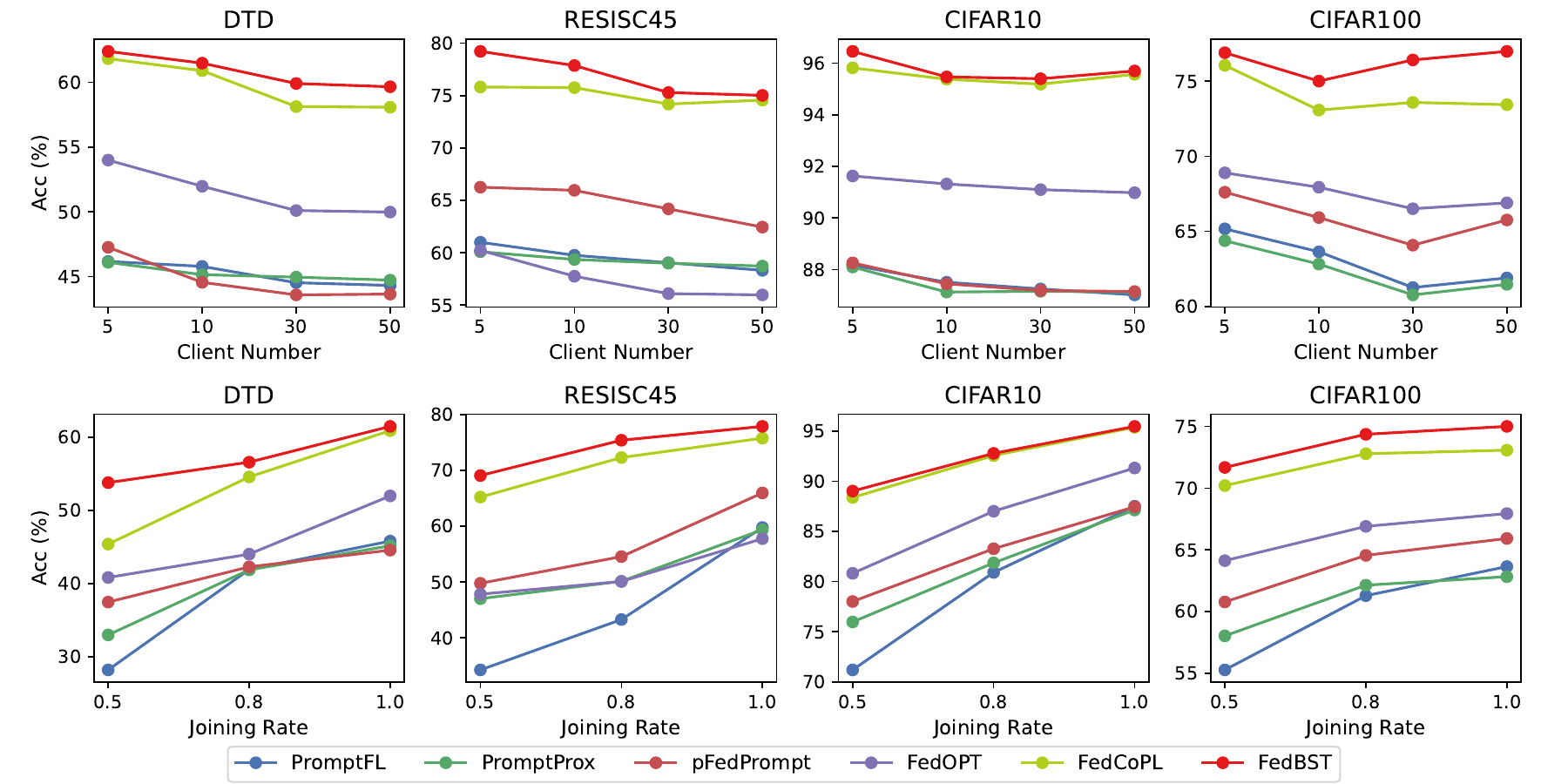}
  \vspace{-8pt}
  \caption{Results of experiments with various client numbers and different client joining rates under Dirichlet-based label skews ($\beta=0.1$).}
  \label{fig: client}
  \vspace{-15pt}
\end{figure*}

\subsection{Experimental Results}
\noindent \textbf{Results under Dirichlet-based label skew with various datasets.}
Table~\ref{tab: accuracy Dirichlet} reports the accuracy under Dirichlet-based skew with $\beta \in \{0.1, 0.05\}$.
FedBST consistently achieves the best performance across all settings.
Notably, on fine-grained datasets like CUB, baselines relying on FPL or CPL often succumb to noisy pseudo-labels, leading to error accumulation and occasionally underperforming zero-shot CLIP.
While FedCoPL provides a strong unsupervised federated classification baseline, FedBST yields further consistent improvements.
This indicates that effective utilization of ambiguous samples and the visual-text ensemble is critical for robust UFC with CLIP.

\noindent \textbf{Results under quantity-based label skew with various datasets.}
Table~\ref{tab: accuracy Quantity} presents results where clients observe only a limited subset of classes ($s = C \times 0.2$).
This severe class scarcity significantly challenges the generation of reliable pseudo-labels, causing several baselines to underperform zero-shot CLIP.
In contrast, FedBST consistently achieves the best results, outperforming the strongest baseline FedCoPL by a clear margin (Average +0.95\%).
This demonstrates that our bilateral self-training strategy effectively mines useful signals from ``waist'' samples even when high-confidence data is scarce, ensuring robustness under extreme non-IID conditions.

\begin{table}[t]
\centering
\caption{Experimental results (\%) using CLIP ViT-B/16 as backbone under Dirichlet-based label skew ($\beta=0.1$).}
\label{tab: accuracy backbone B16}
  \vspace{-6pt}
  \renewcommand{\arraystretch}{1.1}
\resizebox{0.45\textwidth}{!}{ 
\begin{tabular}{lccccc}
\toprule
{Method} & \textbf{PL} & {{DTD}} & {{RESISC45}} & {{CUB}} & {{UCF101}}  \\ \midrule
CLIP & - & 42.87 & 56.61 & 55.16 & 65.13 \\
PromptFL~\cite{guo2023promptfl} & FPL & 44.36 & 61.36 & 51.95 & 64.60 \\
PromptProx~\cite{li2020federated} & FPL & 47.12 & 61.01 & 49.44 & 65.31 \\
pFedPrompt~\cite{guo2023pfedprompt} & FPL & 47.65 & 63.18 & 50.19 & 65.19 \\
FedOPT~\cite{li2024global} & FPL  & 46.52 & 58.94 & 49.67 & 65.68  \\ 
FedCoPL~\cite{guo2025cooperative} & CoPL & 56.83 & 77.50 & 59.78 & 78.29 \\ \rowcolor[gray]{0.9}
FedBST & UACP &  \textbf{58.72} & \textbf{79.81} & \textbf{61.40} & \textbf{78.93} \\ 	 	 	 
\bottomrule
\end{tabular}
}
\vspace{-12pt}
\end{table}

\subsection{Analysis}

\noindent \textbf{Results under different client numbers.}
Fig.~\ref{fig: client} (Top) illustrates the performance under varying federation scales.
As client numbers increase, data heterogeneity intensifies, causing performance degradation across all methods due to the difficulty of reliable pseudo-labeling.
However, FedBST exhibits the slowest degradation rate and maintains the highest accuracy.
This demonstrates its superior scalability, attributed to the UACP module, which effectively prevents noise accumulation even when local data becomes sparse.

\begin{table}[t]
\centering
\caption{Experimental results (\%) using CLIP RN50 as base model under Dirichlet-based label skews ($\beta=0.1$) across four datasets.}
\vspace{-6pt}
\label{tab: accuracy backbone RN50}
\renewcommand{\arraystretch}{1.1}
\resizebox{0.45\textwidth}{!}{ 
\begin{tabular}{lccccc}
\toprule
{Method} & \textbf{PL} & {{DTD}} & {{RESISC45}}  & {{UCF101}} & {{CIFAR10}} \\ \midrule
CLIP & - & 41.62 & 52.12  & 65.13 & 74.80\\
PromptFL~\cite{guo2023promptfl} & FPL & 42.36 & 56.16  & 62.90 & 78.01\\
PromptProx~\cite{li2020federated} & FPL & 43.12 & 56.54  & 64.37 & 78.77\\
pFedPrompt~\cite{guo2023pfedprompt} & FPL & 43.65 & 57.13  & 64.97 & 79.69\\
FedOPT~\cite{li2024global} & FPL  & 45.52 & 58.94  & 65.54 & 80.13 \\ 
FedCoPL~\cite{guo2025cooperative} & CoPL & 51.83 & 70.78  & 76.48 & 84.85 \\ \rowcolor[gray]{0.9}
FedBST & UACP & \textbf{53.40} & \textbf{72.55} & \textbf{77.01} & \textbf{85.21} \\
\bottomrule
\end{tabular}
}
\vspace{-18pt}
\end{table}

\noindent \textbf{Impact of client joining rates.}
Fig.~\ref{fig: client} (Bottom) reports results under varying participation rates.
Lower participation reduces information aggregation and increases local update variance, causing performance drops across all methods.
Baselines are particularly sensitive to this scarcity, showing substantial degradation on challenging datasets.
In contrast, FedBST consistently outperforms competitors, demonstrating that our visual-text ensemble strategy ensures stability even when active clients are sparse.

\noindent \textbf{Results under different image encoder backbones.}
Tables~\ref{tab: accuracy backbone B16} and~\ref{tab: accuracy backbone RN50} present results under Dirichlet-based label skews using ViT-B/16 and RN50 image encoders.
FedBST consistently outperforms zero-shot CLIP and federated baselines across both Transformer-based and CNN-based architectures.
This confirms that our framework is architecture-agnostic, maintaining superior performance and practical applicability without requiring backbone-specific modifications.

\begin{table}[!t]
\centering
\caption{Performance (\%) of FedBST under different values of hyperparameter $\tau_{con}$ and $\lambda$ with Dirichlet-based label skews.}
\vspace{-4pt}
\label{tab: hyperparameter 1}
\resizebox{.48\textwidth}{!}{ 
\begin{tabular}{lccccccc}
\toprule
$\lambda$ ($\times e^{-3}$) &  0.4 & 0.6 & 0.8 & 1.0 & 1.2 & 1.4 & 1.6 \\ \midrule
DTD & 60.24 & 60.10 & 60.71 & \textbf{61.46} & 60.73 & 60.57 & 60.79   \\
RESICS45 & \textbf{78.23} & 78.01 & 78.20 & 77.88 & 77.60 & 77.91 & 78.05  \\ \midrule
$\alpha$ & 0.60  & 0.65 & 0.70 & 0.75 & 0.80 & 0.85 & 0.90 \\ \midrule
DTD & 61.31 & 61.26 & 61.15 & 61.31 & \textbf{61.46} & 61.36 & 61.19   \\
RESICS45 & \textbf{78.08} & 77.97 & 77.91 & 77.93& 77.88 & 77.75  & 77.71  \\ 
\bottomrule
\end{tabular}
}
\vspace{-12pt}
\end{table}

\begin{table}[!t]
\centering
\caption{Performance (\%) of FedBST under different values of hyperparameter $\tau_{con}$ and $\tau_{ent}$.}
\vspace{-4pt}
\label{tab: hyperparameter 2}
\renewcommand{\arraystretch}{0.9}
\resizebox{.38\textwidth}{!}{ 
\begin{tabular}{lccccccc}
\toprule
$\tau_{con}$ & 0.10 & 0.15 & 0.20 & 0.25 & 0.30 \\ \midrule
DTD & 61.21 & 61.45 & 61.46 & \textbf{61.62} & 61.60    \\ 	
RESICS45 & \textbf{78.56} & 78.10 & 77.88 & 77.39 & 77.30  \\   	\midrule 
$\tau_{ent}$ & 0.40 & 0.45 & 0.50 & 0.55 & 0.60 \\ \midrule
DTD &  61.39 & 61.50 & 61.46 & \textbf{61.55} & 61.21 \\ 	
RESICS45 & \textbf{78.33} & 78.04 & 77.88 & 77.39 & 76.87  \\  	
\bottomrule
\end{tabular}
}
\vspace{-18pt}
\end{table}

\begin{table}[t]
\centering
\setlength{\abovecaptionskip}{0.1cm}
\caption{Ablation results (\%) comparing the effects of Uncertainty-Aware Confidence Penalization (\textbf{UACP}) and Visual-Text Ensemble (\textbf{VTE}). }
\vspace{-4pt}
\label{tab: ablation_partial}{}
\resizebox{.43\textwidth}{!}{ 
\begin{tabular}{cccccc}
\toprule
\textbf{UACP} & \textbf{VTE} & DTD & RESISC45 & CUB & CIFAR100\\ \midrule
- & - &  60.89 & 75.76 & 56.09 & 73.59 \\
\cmark & -  &  61.22 & 77.34 & 56.89 & 74.87  \\
- & \cmark &  61.10 & 76.77 & 56.54 & 74.41 \\
\cmark & \cmark &  61.46 & 77.88 & 57.25 & 75.01  \\
\bottomrule
\end{tabular}
}
\vspace{-18pt}
\end{table}

\noindent \textbf{Parameter Sensitivity.} 

Tables~\ref{tab: hyperparameter 1} and~\ref{tab: hyperparameter 2} report performance under varying $\lambda$, $\alpha$, $\tau_{\text{con}}$, and $\tau_{\text{ent}}$.
FedBST exhibits stable performance across a broad range of values, with accuracy fluctuating minimally on DTD and RESISC45 around the default settings.
This resilience suggests that FedBST is practical for real-world scenarios where extensive hyperparameter tuning is infeasible.

\noindent \textbf{Effectiveness of each component.}
Table~\ref{tab: ablation_partial} presents an ablation study on Uncertainty-Aware Confidence Penalization (UACP) and Visual-Text Ensemble (VTE) under Dirichlet-based label skew ($\beta = 0.1$). 
Each component provides individual gains, yet their combination performs best. This confirms their synergy: UACP regularizes training, while VTE calibrates inference.

\section{Conclusion}
This work introduces FedBST, a bilateral self-training framework for unsupervised federated classification that is explicitly designed to operate under severe non-IID conditions.
Rather than discarding ambiguous data, FedBST makes systematic use of both high-confidence and uncertain samples, allowing a more complete exploitation of local data.
Uncertain predictions are incorporated as regularization signals via uncertainty-aware confidence penalization, while inference is further strengthened through a visual-text ensemble, which together help alleviate confirmation bias and improve the alignment between global semantics and client-specific distributions.
Experimental results across multiple benchmarks indicate that these design choices lead to stable performance improvements, with the largest gains observed in settings characterized by strong label skew and limited data.

\section*{Acknowledgment}

This work was supported by the National Natural Science Foundation of China (NSFC) (Grant 62376265).
It was also supported by the National Natural Science Foundation of China under Grants (62276256, U2441251).

\bibliographystyle{IEEEbib}
\bibliography{icme2026references}

\end{document}